# Delegating via Quitting Games

Juan Afanador[*,1], Nir Oren[1], and Murilo S. Baptista[2]

[1]Department of Computer Science, University of Aberdeen
[2]Institute of Complex Sciences and Mathematical Biology, University of Aberdeen

April 20, 2018

## Abstract

Delegation allows an agent to request that another agent completes a task. In many situations the task may be delegated onwards, and this process can repeat until it is eventually, successfully or unsuccessfully, performed. We consider policies to guide an agent in choosing who to delegate to when such recursive interactions are possible. These policies, based on quitting games and multi-armed bandits, were empirically tested for effectiveness. Our results indicate that the quitting game based policies outperform those which do not explicitly account for the recursive nature of delegation.

## 1 Introduction

Task delegation is a common feature of multi-agent systems. Agents seeking to increase the likelihood of successful task achievement may decide to hand it over to more experienced agents. For example, an agent representing an user may delegate a transport booking task to a travel agent, as the latter has more expertise in the domain than the former. Much of the research within the field of trust and reputation, e.g. principal-agent theory and multi-armed bandit problems, has considered the problem of who to delegate to when maximising the likelihood of success, how to price such delegation requests, and how to learn from them. However, to our knowledge, nearly all such work has considered a single instance of delegation [7, 1, 14, 20]. That is, a delegation request is generated from one agent to another, whence the process is terminated.

In this paper, we consider consecutive requests reaching some agent at the end of a *delegation chain*, who finally executes the task. We show that in an environment where the competency of others is initially unknown or partially known, existing approaches –concerned with a single level of delegation– are not optimal. We analyse the problem of recursive task delegation through the lens of game theory, formulating a quitting game mechanism that allows for either consecutive delegation, or immediate execution.

We demonstrate empirically that such a game theoretic approach outperforms more traditional strategies, namely $\epsilon-$greedy, UCB1, and Gittins Index guided partner selection. The main contributions of this paper rest on the formal representation of recursive delegation as both a quitting game and a multi-armed bandit problem, along with the generation of empirical evidence on their relative performance. In particular, it is shown that the quitting game surpasses existing techniques, in terms of the speed of convergence and the quality of the overall solution.

---

[*]r01jca16@abdn.ac.uk

The remainder of this paper is structured as follows. In the next section, the Gittins Index, Upper Confidence Bound algorithms, and quitting games are introduced. Section 3 presents our principal contribution: the recursive delegation problem is formalised by considering agents capable of delegating along a tree, and the implementation of the corresponding quitting game explained. We proceed to evaluate our approach in Section 4. Section 5 concludes the analysis with a discussion of our results and related work.

## 2 Background: Optimal Indexing, Optimism, and Quitting Games

We begin this section by describing the multi-armed bandit problem and its solution through the application of the Gittins Index [10]. Multi-armed bandit problems are often used to reason about how exploration and exploitation of resources should take place in an uncertain or unknown environment. In the context of delegation they rationalise the selection of the agent to be entrusted with the task. Following this, we describe UCB algorithms and quitting games.

### 2.1 Multi-Armed Bandits

Sequential decision problems are usually represented as multi-armed bandits (MABs). A single agent repeatedly selects one among finitely many actions, obtaining a stream of rewards subject to an unknown stochastic process. Rewards are additive and discounted in time.

**Definition 2.1** (**Multi-Armed Bandit Problem** [19]). A Multi-armed bandit is a tuple $M = \langle (A_i, x_i, r_i, f_i)_{i \in \mathcal{N}}, \boldsymbol{w} \rangle$. A single agent selects one state from $A_i = \{inactive, active\} \mapsto \{0, 1\}$ affecting one arm $x_i$ among $\mathcal{N} = \{0, \ldots, N\}$, during $T$ trials. $r_i$ and $\boldsymbol{w}$ denote the reward attached to the ith arm, and a vector of independent random variables. The agent's decision, for a time-discounting parameter $\beta \in [0, 1]$, is given by the following elements:

**Local Time (of an Arm):** $n_i(t), t = \{0, 1, \ldots, T\}$. Number of times an arm is operated.

**Arm:** $\{x_i(n_i(t)), r_i(x_i(n_i(t))); n_i(t) = \{0, 1, \ldots, t\}\}$. Arms are fully characterised by their state $(x_i(\cdot))$ and reward $(r_i(\cdot))$.

**Policy:** A rule $\gamma_t$ dictating the current control action $u(t)$, such that:

$$u(t) \equiv \gamma_t : \times_{i=1}^{N} z_i \times_{m=1}^{t-1} u(m) \to \bigcup_{i=1}^{N} e_i$$

$$\text{where:} \quad e_j = (\underbrace{0, 0, \ldots, 1, \ldots, 0}_{\text{j arms}})^{\overbrace{\phantom{0,0,...,1,...,0}}^{\text{N arms}}},$$

$$z_i(t) := [x_i(0), x_i(1), \ldots, x_i(t)].$$

**Evolution Rules (ER):**

$$x_i(n_i(t+1)) = \begin{cases} f_{n_i(t)}(x_i(0), \ldots, x_i(n_i(t)), w_i(n_i(t))) & , u_i(t) = 1 \\ x_i(n_i(t)) & , u_i(t) = 0 \end{cases}$$

where:

$$n_i(t+1) = \begin{cases} n_i(t) + 1 &, u_i(t) = 1 \\ n_i(t) &, u_i(t) = 0 \end{cases},$$

$$u_i(t)) = \begin{cases} r_i(x(n_i(t+1))) &, u_i(t) = 1 \\ 0 &, u_i(t) = 0 \end{cases}.$$

**Agent's Optimisation Problem:**

$$V(\gamma, T) := max_\gamma E[\sum_{t=0}^{T} \beta^t \sum_{i=1}^{N} r_i(x_i(n_i(t)), u_i(t)) \mid z(0)] \; s.t. \; ER.$$

$\square$

The Gittins Index solves MABs optimally. It assigns each arm a dynamically calculated value. During each trail, the optimal strategy amounts to selecting the highest valued arm.

**Proposition 2.1** (**Gittins Index** [28]). *There exist functions:*

$$G_i(z_i(t)) = \sup_{\sigma > 0} \frac{E[\sum_{t=0}^{\sigma-1} \beta^t r_i(x_i(t)) \mid z_i(0) = z_i]}{E[\sum_{t=0}^{\sigma-1} \beta^t \mid z_i(0) = z_i]}, \forall i \in N$$

*such that a policy $\gamma^*$ activating the arm*

$$x_n, \quad n \in \underset{m \in \{0, \ldots, N-1\}}{argmax} \{G_m(z_m(t))\}$$

*is optimal, for a time-discounting parameter $\beta \in [0, 1]$. The function $G_m(\cdot)$ is calculated from the dynamics of the stochastic process alone. The function and value it returns is referred to as the* Gittins Index. $\square$

Alternative approaches rely on heuristics or additional variables such as regret (the difference between the sum of the best possible outcome and the sum of actual rewards), risk (the variance of the expected reward), or optimism (the value of the upper confidence bounds on the expected rewards):

**UCB:** Estimate a ceiling for each arm's reward, updating the resulting Upper Confidence Bound (UCB) after every realisation. UCB1 guarantees asymptotically logarithmic regret [2].

**$\epsilon$-Greedy:** Sweeps through the available set of arms before selecting the most rewarding with probability $\epsilon$, shifting to a random arm with probability $1 - \epsilon$ [9].

**Risk-Averse:** Require some previous knowledge on the value of the optimal arm, as to achieve a randomised policy that attains uniformly bounded risk over finite time [25].

Variations of these give rise to the algorithms employed in different versions of MABs, from superprocesses [6] to risk-averse bandits [25], via contextual [18], arm-acquiring [29] and restless [3] bandits. Non-frequentist bandits also use UCB on the posterior knowledge of rewards to guide the exploratory stage of the process [21].

## 2.2 Upper Confidence Bound Algorithms

Regret, or the gap between the cumulative expected reward from the best arm available ($\sum_{t \in \{1,...,T\}} \mu^*$) and the cumulative expected reward induced by the solution of interest ($E[V(\gamma, T)]$), is considered a valid criterion of performance.

Commonly computed as [2]:

$$Regret(\gamma, T) := \mu^* T - E[V(\gamma, T)],$$

it has become the default policy objective for many algorithms, due to its practical and intuitive appealing. The UCB family, as briefly mentioned, is among them.

UCB algorithms are based on the idea that an optimistic guess on the available arms is a good way to achieve minimal regret. Formally, the probability of deviating from the mean expected reward dictates the construction of the upper bound, as prescribed by the Chernoff-Hoeffding inequality [13]. If $\{x_i\}_{i \in N}$ are independent random variables satisfying $P(b_i \leq x_i \leq c_i) = 1$, then for $X := \sum_{k=1}^{N} x_k$, $\mu := E[X]$, and all $t \in \{0, \ldots, T\}$:

$$P(X - \mu \geq t) \leq e^{-2t^2 / \sum (c_i - b_i)^2}.$$

A sensible choice of parameters recasts the above expression into:

$$P(X - \mu \geq a(t, i)) \leq e^{-2n_i a^2}, a(T, i) = \sqrt{2 Log(T)} / n_i,$$

suggesting the existence of an exponentially decaying upper bound for the mean dispersion of rewards:

$$P(X + a(t, i) \geq \mu) \leq T^{-4}.$$

In contrast to the Gittins Index, which guarantees $\Omega(\sqrt{NT})$ expected regret on MABs, UCB1 is an optimistic algorithm, i.e., it aims at the maximisation of $a(\cdot)$, known to achieve $O(\sqrt{NT \log T})$ regret [11]. Its "delegational" interpretation is illustrated in Algorithm 1.

UCB1 also secures a theoretical upper bound on regret:

**Proposition 2.2** (**UCB1's Theoretical Upper Bound** [2]). *UCB1 attains regret that is at most:*

$$8 \sum_{i : \mu_i < \mu^*} \frac{log(T)}{\delta_i} + \left(1 + \frac{\pi^2}{3}\right) \sum_{j=1}^{K} \delta_j$$

*where* $\delta_i := \mu * - \mu_i$. □

Solutions akin to an optimistic version of the Gittins Index offer a slightly different view on the bounds of optimal regret minimisation for standard MABs. That is, if the notion of a retirement value is invoked as to facilitate the index's computation [11], it can be demonstrated that:

**Proposition 2.3** (**Optimistic Gittins Theoretical Bound** [11]). *An Optimistic Gittins solution for MABs, achieves an upper bound on regret given by:*

$$Regret(\gamma, T; \theta) \leq \sum_{i \neq i^*} \frac{(1 + \epsilon)(\theta^* - \theta_i)}{d(\theta^*, \theta_i)} log(T) + C(\epsilon, \theta)$$

*where* $\theta_j$ *denote the success probability per trial, parametrising the prior distribution over* $\{x_i\}_{i \in N}$. □

This result not only bridges UCB heuristics with a principled approach to MABs, but also establishes lower bounds for regret, matching Lai and Robbins' [17]. In fact, it has been proved that a greedy strategy with respect to the simplest of those optimistic approximations to the Gittins Index, agrees with Proposition 2.2.

## Algorithm 1 UCB1 Delegation

**Input:** $\{a_i, ad_i\}_{i \in N}$: Tuples of agents and their neighbours,
  $K$: Number of neighbours per agent.
**Output:** $S$: Sequence of agents to whom its has been delegated to, $\mu$: Array of rewards.
2: **procedure** UCB1$(a; K)$
  $S \leftarrow \emptyset, \mu \leftarrow \{\mu_i\}_{i \in N}, \mu_i \sim Beta(1,1)$.
4: **for** j=1$\rightarrow N$ **do**
    $ad_i \leftarrow \{a_j\}_{j \neq i}, CountSuccess_{a_{(\cdot)_j}} \leftarrow 0$
6:   $\alpha_j = CountSuccess_{a_j}, \beta_j = CountSuccess_{a_k \in ad_j}$
    $\mu_j \leftarrow \frac{1}{1+\beta_j/\alpha_j}$
8:   Delegate to $a_m = argmax(\mu_m + \sqrt{2log(trials)}/(\alpha_m + \beta_m)$
    $S \leftarrow S \cup \{a_m\}$
10:  **if** $Outcome == True$ **then**
     $CountSuccess_{a_i} \leftarrow CountSuccess_{a_i} + 1$
12:  **else**
     $CountSuccess_{a_m} \leftarrow CountSuccess_{a_m} + 1$
14: **return** $Outcome$

The superior empirical performance and lighter computational burden of the UCB family, to which its popularity is attributed, can also be attained through algorithms engendered by directly tackling the analytical formulation of MABs. Consequently, it seems reasonable to expect that game-theoretic approaches, relying on an explicit representation of the actual interaction among different agents, have the potential to bring about greater accuracy not necessarily penalised by lack of practicality.

### 2.3 Quitting Games

Stochastic games with embedded actions characterise both sequential and recursive problems. They bring forth the importance of agency through strategic interaction. Quitting games, a sub-class of stochastic games, provide a good example.

A quitting game (QG) presents agents with continuing ($c$) or quitting ($q$) as the only possible actions. It is framed in a finite time horizon and the rewards are collected whenever $q$ is played or the game's terminal time is reached.

**Definition 2.2 (Quitting Game).** A quitting game is a pair $(\mathcal{N}, \{r_s\}_{S \subseteq \mathcal{N}})$, where $\mathcal{N} = \{1, \ldots, N\}$ indicates the number of players, and $r_S$ the rewards obtained once a subset $S \subseteq \mathcal{N}$ quits the game.

**Actions:** $A_i = \{c^i, q^i\}, \forall i \in \mathcal{N}$.

**Strategy:** $(x_t^i) : T \to [0,1], \forall i \in \mathcal{N}$, where $x_t^i$ is the probability of the ith player playing $c_i$ at trial $t \in T \subseteq \mathbb{N}^+$.

**Profile:** $\mathbf{x} := (x_t)_{t \in T}$. It induces a probability distribution $\mathbf{P_x}$ over the set of actions. $\mathbf{E_x}$ denotes the corresponding expectation operator.

**Expected Pay-off:** $v_i(\mathbf{x}) := \mathbf{E_x}[r_{S_\tau} I_{\tau < \infty}], \forall i \in N$, for terminating time $\tau = inf\{t, S_t \neq \emptyset\}$.

□

For two agents the game in normal form appears in Table 1. The first entry in each pair corresponds to $Agent_0$'s pay-off, the other is $Agent_1$'s. Thus, for example, whenever $(c_0, q_1)$ is played, $Agent_0$ receives $b_0$ and $Agent_1$ gets $b_1$. $(c_0, c_1)$ continues the game leading to yet unrealised rewards, this situation is represented by "↻".

Table 1: Normal Form of a Quitting Game

|  |  | $Agent_1$ |  |
|---|---|---|---|
|  |  | $c_1$ | $q_1$ |
| $Agent_0$ | $c_0$ | ↻ | ( $b_0$ , $b_1$ ) |
|  | $q_0$ | ($a_0$ , $a_1$) | ( $d_0$ , $d_1$ ) |

For $a_0 > 0$, $a_1 < d_1$, $d_0 < b_0$, and $a_1 \geq b_1$, the stationary profile $(\mathbf{x}^0, c_1)$, $x_t^0 \ll 1$ is an $\epsilon - equilibrium$. More generally, in [23] it is proved that for every quitting game where players prefer unilateral termination to indefinite continuation there exists a cyclic subgame perfect $\epsilon - equilibrium$; and in [24] it is shown that two and three players QG's have an stationary $\epsilon - equilibrium$ of the kind just presented.

**Proposition 2.4** ($\epsilon - equilibria$ in **Quitting Games** [24]). *Let $\epsilon > 0$. Any game satisfying: $r^i_{\{i\}} = 1, \forall i \in \mathcal{N}$; and $r^i_S \leq 1, \forall i \in S$; has a cyclic subgame perfect $\epsilon - equilibrium$.* □

Repeated delegation might be sub-optimal even for a non-binding time restriction, were it feasible to map the decision problem directly onto a QG. It is not only a matter of selecting the fittest agent available for the task, but also finding out whether those higher up in the delegation chain could have achieved a better expected outcome by delegating to others. The resemblance to the exploration/exploitation-dilemma in MABs is more than apparent.

The Gittins Index, UCB1 and quitting games reflect different features of delegation. The Gittins Index synthesises the multidimensionality of the dynamic optimisation problem behind decision making. UCB1 provides a practical alternative to the computationally demanding calculation of the index; while game-theoretical models assert the strategic aspect of delegation by exploiting potentially conflicting objectives. To build upon these observations a delegation problem in the form of a quitting game will be evaluated using three different extensions of multi-armed bandits as benchmarks.

## 3 Approach: Delegation and Games

MABs provide an appropriate framework for modelling delegation, at certain risks. Sen et al. (2015), for instance, evaluated algorithms drawn from the greedy family over different tree-like networks, running on a standard Gaussian process. The variable of interest was the cost-benefit ratio of the arm activation.

The high rewarding task executioners were identified, employing budgetary restrictions to probe for reliability, while agents' motives remained independent of one another as if there were no actual handing-over of a task, but the mere repetition of a discarding process. However the dependency of those who have been delegated to, as potential agents of delegation, on the latter's already constituted objectives should be the driving force behind the "delegational" interpretations of MABs. It appears that such dependency lies at the heart of its recursive nature.

Given this we opted for a dynamic indexing adaptation of MABs that used the Gittins Index. Because of its analytical derivation and optimal properties, exclusively based on the history of past delegations, the decision seemed consequential.

Noting that decisions can be associated with a binary random variable keeping track of successful and failed choices, a Bernoulli process is a good description of monitoring behaviour throughout delegation. For a large number of trials ($N$), and a parameter ($\delta$), discounting future rewards at a rate $c$, i.e., $\delta = e^{-c}$, in $(0.8, 1]$, the Gittins Index may be approximated with a closed-form function [5]:

$$G(N) \approx \mu + \sqrt{\frac{\mu(1-\mu)}{N+1}} \psi\left(\frac{1}{(N+1)^c}\right)$$

$\mu$ is the mean of the compound (Beta) distribution of the random variable indicating a successful delegation, and

$$\psi(\tau) = \begin{cases} \sqrt{\tau/2} & , \tau \leq 0.2 \\ 0.49 - (0.11\tau)^{-1/2} & , \tau \in (0.2, 1] \\ 0.63 - (0.26\tau)^{-1/2} & , \tau \in (1, 5] \\ 0.77 - (0.58\tau)^{-1/2} & , \tau \in (5, 15] \\ \{2log(\tau) - loglog(\tau) - log(16\pi)\}^{1/2} & , otherwise \end{cases}$$

is, in turn, the approximation to the boundary of the continuation region $\mathcal{C} = \{(z, s) : z > -b(s)\}$ for an optimal policy $\gamma^*$ dependent on the stopping time $\tau$, in accordance with the notation in Definition 2.1 and Proposition 2.1. Algorithm 2 (DID) summarises all these elements.

---

**Algorithm 2** Dynamically Indexed Delegation (DID)

---

**Input:** $A := \{a_i, ad_i\}_{i \in N}$: Tuples of agents and their neighbours,
$K$: Number of neighbours per agent, $\delta$: Time-discounting parameter.
**Output:** $S$: Sequence of agents receiving a delegation request, $\mu$: Array of rewards.
2: **procedure** DL($A; K$)
    $S \leftarrow \emptyset, \mu \leftarrow \{\mu_i\}_{i \in N}, \mu_i \sim Beta(1,1)$.
4:    **for** j=1$\rightarrow N$ **do**
       $ad_i \leftarrow \{a_j\}_{j \neq i \in K}, CountSuccess_{a_j} \leftarrow 0, \delta_j \leftarrow [0.8, 1)$
6:       $\alpha_j = CountSuccess_{a_j}, \beta_j = CountSuccess_{a_k \in ad_j}$
       $\mu_j \leftarrow \frac{1}{(1+\beta_j/\alpha_j))}$
8:       $G_i \leftarrow \mu_j + (\frac{\mu_j(1-\mu_j)}{\alpha_j+\beta_j+1})^{1/2} \psi\left(1/(\alpha_j + \beta_j + 1)log(\delta_i^{-1})\right)$
       Delegate to $a_m = argmax(\{G_k\}_{k \in ad_i})$
10:     $S \leftarrow S \cup \{a_m\}$
       **if** $Outcome == True$ **then**
12:       $CountSuccess_{a_i} \leftarrow CountSuccess_{a_i} + 1$
       **else**
14:       $CountSuccess_{a_m} \leftarrow CountSuccess_{a_m} + 1$
    **return** $Outcome$

---

A standard greedy algorithm, following the description in Section 2 was also considered an appropriate accompanying benchmark, as it is not yet clear whether the algebraic and topological characteristics of a delegation problem allowed for greediness to be optimal. Along with the MAB extension, they will both serve as a point of reference for QG's. Algorithm 3 ($\epsilon$-greedy) gives a full characterisation.

QG's on the other hand seem more than fitting for modelling the latency of sequential subdelegation. They have been explicitly conceived for representing the opening of new instances of the problem that gives rise to the whole game. Thus delegation problems will be cast as delegation games.

**Definition 3.1 (Delegation Game).** A Delegation Game is characterised by the tuple $G = \langle N, (A^i, U^i, r^i)_{i \in \mathcal{N}}, \mathbf{x} \rangle$, where every agent in $\mathcal{N} = \{1, \ldots, N\}$ has the following attributes:

**Actions:** $A_t^i := \{d_t^i, e_t^i\}, \forall i \in \mathcal{N}$. $\Delta(A_i)$ is the set of all probability distributions over the set of agent i's actions.

**Startegy:** $(x_t^i) : A_i \rightarrow [0, 1], \forall i \in \mathcal{N}$, where $x_t^i$ is the probability of the $i$th player playing $d_t^i$.

**Profile:** $\mathbf{x}^i := (x_t^i)_{t \in T}$. It induces a probability distribution $\mathbf{P_x} \in \Delta(A)$ over the set of actions.

**Rewards:** $r_t^i : \times_{j \in \mathcal{N}} \Delta(A^j) \times D_t \to \mathbb{R}, \forall i \in \mathcal{N}$ is a Lebesgue measurable function representing the gains from delegation for each player at stage $t \in T \subseteq \mathbb{N}^+$ when a set of agents $D_t$ have been delegated to.

**Updating Rule:** $u_i : A_{t-1} \times \mathbb{R} \to \Delta(A^i)$ is a measurable set-valued function that dictates the transition from one state of the system to a potentially different profile.

□

The Delegation Game falls into one of three states: continuous delegation ($s_c$), immediate execution of the task by another agent ($s_m$), or self-execution of the task ($s_e$). The last two are absorbing since no agent can leave once they are reached.

Table 2: Normal Form of a Delegation Game

| | $e_1$ |
|---|---|
| $e_0$ | $(Y_0, 0)$ |

$s_e$

| | $d_1$ | $e_1$ |
|---|---|---|
| $d_0$ | ↻ | $(X_0, Y_1)$ |
| $e_0$ | $(Y_0, 0)$ | $(Y_0, 0)$ |

$s_c$

| | $d_1$ |
|---|---|
| $d_0$ | $(X_0, Y_1)$ |

$s_m$

The agent receiving the task is chosen by comparing the discounted expected values at $t$, attached to the decision of whether delegating or not i.e.,

$$v_i(\mathbf{x}^*) \geq v_i(\mathbf{x}^i, \mathbf{x}^{-i}), \forall \mathbf{x}^i \in \Delta(A^i)$$

It is a more robust solution concept than the one in Proposition 2.4. This approach to its empirical evaluation would provide evidence of the potential optimality of the game-theoretic approach to delegation, and also induce the exploration of Nash equilibria or related refinements for this sub-class of games. Algorithm 4 describes its application to Definition 3.1.

---

**Algorithm 3** $\epsilon$-Greedy Delegation

---

**Input:** $A := \{a_i, ad_i\}_{i \in N}$: Tuples of agents and their neighbours, $\epsilon$: Array of greedy parameters, $K$: Number of neighbours per agent.
**Output:** $S$: Sequence of agents receiving a delegation request,
$\mu$: Array of rewards.
2: **procedure** ED($A; K$)
    $S \leftarrow \emptyset, \mu \leftarrow \{\mu_i\}_{i \in N}, \mu_i \sim Beta(1, 1)$.
4:    **for** j=1→ N **do**
        $ad_i \leftarrow \{a_j\}_{j \neq i \in B}, CountSuccess_{a_j} \leftarrow 0, \delta_j \leftarrow (0, 1), \epsilon_j \leftarrow (0.05, 0.1)$
6:        $\alpha_j = CountSuccess_{a_j}, \beta_j = CountSuccess_{a_k \in adj}$
        $\mu_j \leftarrow \frac{1}{1 + \beta_j / \alpha_j}$
8:    **if** $random() < \epsilon_j$ **then**
            Delegate to $a_m$ s.t. $\exists m \neq i, m \in ad_i$
10:    **else**
            Delegate to $a_m = argmax(\{\mu_k\}_{k \in ad_i})$
12:        $S \leftarrow S \cup \{a_m\}$
    **if** $Outcome == True$ **then**
14:        $CountSuccess_{a_i} \leftarrow CountSuccess_{a_i} + 1$
    **else**
16:        $CountSuccess_{a_m} \leftarrow CountSuccess_{a_m} + 1$
    **return** $Outcome$

---

In Algorithm 4, a set of $K$ neighbours is assigned to each agent, and their respective rewards sampled from an uniform distribution. The resulting initial state allows the computation of individual mixed strategies,i.e., the probabilities of delegating, whenever pairs of agents and neighbours engage in a delegation request. As long as there are neighbours who have not received such a request, despite holding a positive probability of delegating, the selection of the one with the highest expected pay-off will take place. If capable of executing the task, as given by a random "state of nature", this latter agent will have to weigh up the possibility of passing the task down the delegation chain or attempt its completion, thereby triggering a learning process.

---

**Algorithm 4** Delegation Game (DIG)

---

**Input:** $\{a_i, ad_i\}_{i \in N}$: Tuple of agents and their neighbours,
$K$: Number of neighbours per agent, $r$: Array of sampled rewards.
**Output:** $S$: Sequence of agents receiving a delegation request, $x$: Array of mixed strategies.
2: **procedure** DELEGATE($x, r; K$)
    $S \leftarrow \{S_i\}_{i \in N}, x \leftarrow \{x_i\}_{i \in N}, \mu \leftarrow \{\mu_i\}_{i \in M}, r \leftarrow \{r_i\}_{i \in N}$
4:     **for** j=1$\to N$ **do**
        $ad_j \leftarrow \{a_k\}_{k \neq j \in B}, r_j \leftarrow \{\mathcal{U}(r_{j,0}, r_{j,T})\}_{j \in B}, S_j \leftarrow \emptyset, x_j \leftarrow 0$
6:         **for** $a_k \in ad_j$ **do**
            $x_{j,k} = \frac{r_{j,1} - r_{k,0}}{r_{j,2} - r_{k,0}}$
8:             $x \leftarrow x \cup \{x_{j,k}\}$
        **while** $(x \neq \emptyset) \wedge (\exists j[S_j == \emptyset])$ **do**
10:             $m = argmax_{j \in B}(r)$
            **if** $(random() < x_{j,m})$ **then**
12:                 Delegate to $a_m$
                **if** $a_m \in S_j$ **then**
14:                     Update $x_{j,m}, r_{j,m}$
                **else** $a_m \notin S_j$
16:                     $S_j \leftarrow S_j \cup \{a_m\}$
                    **return** LEARN($x_m, r_m; K, a_j$)
18:             **else**
                $a_j$ executes the task
20:                 $S_j \leftarrow \emptyset$
    **return** $(S, x)$

---

    **procedure** LEARN($x_k, r_k; K, a_i$)
2:     **if** $r_{k,0} \leq r_{k,1}$ **then**
        $a_k$ executes the task
4:         Update $x_{i,k}, r_{j,m}$
    **else**
6:         **return** DELEGATE($x_k, r_k; K$)

---

## 4 Evaluation

All four implementations consisted of 1000 trials over 100 runs. DID, UCB1, and $\epsilon$-greedy, required networks modelled after the stable state of the system given by Algorithm 4 (DIG): each agent has 5 neighbours, and may reach up to 4 levels of delegation, in correspondence with the game's final state, whose initial delegation chain was unconstrained.

    The parameter ($\epsilon$) in Algorithm 3 took on values in [0.05,0.1], as seems standard in related work [22, 15, 4]. The discount factor in Algorithm 2 (DID) was set to between [0.8,1], as to remain consistent with the closed-form approximation to the Gittins index. The initial probabilities of delegation were sampled from an uninformative Beta distribution.

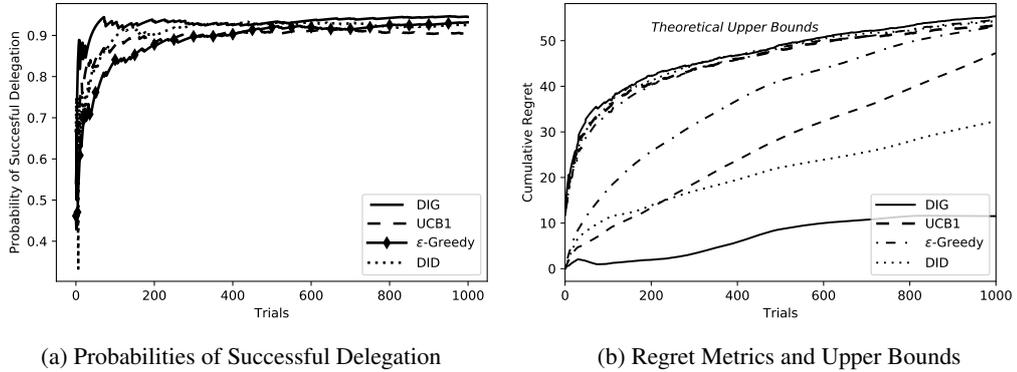

(a) Probabilities of Successful Delegation    (b) Regret Metrics and Upper Bounds

Figure 1: Comparative Performance of the Algorithms

In accordance with Definition 3.1, $d$ ($x$ in Algorithm 4) represents the underlying binomial random variable, keeping track of the probability of delegating. Since it determines the pattern of interaction and, ultimately, the solution to the delegation problem, it is the focus of our attention. The subsequent priors correspond to Beta distributions for an uninformative initial state i.e., $x \sim Beta(\alpha = 1, \beta = 1)$. The posteriors belong to the same family i.e., $x \sim Beta(\alpha + d, \beta + d)$.

Figure 1.a summarises the performance of all three algorithms. $\epsilon$-greedy seems to induce occasional exploratory attempts even after the best solution has been identified, while Gittins struggles to find the right neighbours, given the structure of the problem. The game, on the other hand, finds the optimal strategy rather easily. The continuous quitting of players along different links of the final chain leads to agents with the adequate set of opportunity costs, opening up a path to an optimal solution.

| Algorithm | Final Probability of Successful Execution | Mean Rate of Convergence | Mean Regret |
|---|---|---|---|
| DIG | 0.931 95%CR[0.930,0.933] | 0.441 | 7.124 |
| DID | 0.910 95%CR[0.907,0.913] | 0.410 | 20.939 |
| $\epsilon$-Greedy | 0.898 95%CR[0.894,0.902] | 0.364 | 37.400 |
| UCB1 | 0.896 95%CR[0.893,0.898] | 0.344 | 26.860 |

Table 3: Relative Performance of the Algorithms

Table 3 reports the (posterior) probability of a successful delegation, in the context of each algorithm's credible region, mean rate of convergence[1], and mean regret as defined before. The credible regions indicate that there is a 95% probability of finding the true probability within CR[·,·]. The rate of convergence reflects the speed at which the sequence of probabilities approaches 1.

Figure 1.a shows that DIG and DID display higher probabilities of delegation at faster paces of convergence, whereas the heuristic based policies report statistically indiscernible values of all three criteria, with the exception of Mean Regret as UCB1 is specifically designed to attain optimal regret.

---

[1] The mean rate of convergence was approximated by the error of deviating from a probability of delegating equal to 1 ($e_t$), over the first 175 trials i.e., $q \approx \frac{log(e_{t+1}/e_t)}{log(e_t/e_{t-1})}, t \in \{1, \ldots, 175\}$. The cut-off point was obtained through the Welch method [27].

Figure 1.b provides additional support to previous results. It presents each algortihm's behaviour with respect to the theoretical upper bound given by Proposition 2.2. Despite being a feature of UCB1 only, the upper bound was calculated for every algorithm, in order to have a common point of reference. $\epsilon$-greedy performs quite poorly, barely staying within the bound. UCB1 does fulfill the prediction, but cannot keep up with DID. DIG seems to attain a desirable pattern of delegation, which might also explain a certain tolerance for slight drops in performance conducent to increases in the levels of regret.

# 5  Conclusions

We have shown that a game-theoretical approach to delegation is empirically superior to existing techniques for solving either a single delegation instance or a supply-chain request (as in [22]) modelled after MABs. Delegation does not arise as a direct extension of MABs. It, rather, involves multiple loci of agency in constant interaction.

However different from MABs, delegation does share certain similarities, which has led most of the research in the field towards the study of the Gittins Index, Upper Confidence Bound algorithms, and greedy explorations of the domain. In this paper, UCB1, $\epsilon$-greedy, and a closed-form approximation to the Gittins Index (DID) did not only serve as solution templates; their advantages, analytical and practical ones, were put to the test. More importantly, delegation was formulated in terms of a quitting game, i.e., as a delegation game (DIG). The motivation behind this was to evaluate the performance of an algorithm explicitly conceived for modelling recursive decision-making subject to uncertainty.

The probability of delegating, the mean rate of convergence, and the mean levels of regret were used as evaluation criteria. DIG outperformed the other algorithms. It delivered higher probabilities of delegation, at faster rates with lower regret, even surpassing DID -the analytically proven optimal solution to MABs [9].

We believe that further exploration of the results presented here along the lines of hierarchical reinforcement learning [12], feudal learning [8], and more recent applications of Q-learning [16, 26], have the potential of delivering decisive contributions to AI. Determining DIG's computational complexity, and the formal proof of its optimality in relation with the associated combinatorial problem, are left for future work.

# References

[1] R. Alonso and N. Matouschek. Optimal delegation. *The Review of Economic Studies*, 75(1):259–293, 2008.

[2] P. Auer and P. Fischer. Finite-time Analysis of the Multiarmed Bandit Problem*. *Machine Learning*, 47:235–256, 2002.

[3] D. Bertsimas and J. Niño-Mora. Restless Bandits, Linear Programming Relaxations, and a Primal-Dual Index Heuristic. *Operations Research*, 48(1):80–90, 2000.

[4] D. Bouneffouf, A. Bouzeghoub, and A. L. Gançarski. A contextual-bandit algorithm for mobile context-aware recommender system. In *International Conference on Neural Information Processing*, pages 324–331. Springer, 2012.

[5] M. Brezzi and T. L. Lai. Optimal learning and experimentation in bandit problems. *Journal of Economic Dynamics and Control*, 27(1):87–108, 2002.

[6] D. B. Brown and J. E. Smith. Optimal Sequential Exploration: Bandits, Clairvoyants, and Wildcats. *Operations Research*, 61(3):644–665, 2013.

[7] C. Castelfranchi and R. Falcone. Towards a theory of delegation for agent-based systems. *Robotics and Autonomous Systems*, 24(3-4):141–157, 1998.

[8] P. Dayan and G. E. Hinton. Feudal reinforcement learning. In *Advances in neural information processing systems*, pages 271–278, 1993.

[9] E. Frostig and G. Weiss. Four proofs of Gittins' multiarmed bandit theorem. *Annals of Operations Research*, 2016.

[10] J. C. Gittins. Bandit processes and dynamic allocation indices. *Journal of the Royal Statistical Society. Series B (Methodological)*, pages 148–177, 1979.

[11] E. Gutin and V. Farias. Optimistic gittins indices. In *Advances in Neural Information Processing Systems*, pages 3153–3161, 2016.

[12] B. Hengst. *Hierarchical Reinforcement Learning*, pages 495–502. Springer US, Boston, MA, 2010.

[13] W. Hoeffding. Probability inequalities for sums of bounded random variables. *Journal of the American statistical association*, 58(301):13–30, 1963.

[14] F. Koessler and D. Martimort. Optimal delegation with multi-dimensional decisions. *Journal of Economic Theory*, 147(5):1850–1881, 2012.

[15] V. Kuleshov and D. Precup. Algorithms for multi-armed bandit problems. *arXiv preprint arXiv:1402.6028*, 2014.

[16] T. D. Kulkarni, K. Narasimhan, A. Saeedi, and J. Tenenbaum. Hierarchical deep reinforcement learning: Integrating temporal abstraction and intrinsic motivation. In *Advances in neural information processing systems*, pages 3675–3683, 2016.

[17] T. L. Lai and H. Robbins. Asymptotically efficient adaptive allocation rules. *Advances in applied mathematics*, 6(1):4–22, 1985.

[18] T. Lu, D. Pál, and M. Pál. Contextual multi-armed bandits. *International Conference on Artificial Intelligence and Statistics*, 9:485–492, 2010.

[19] A. Mahajan and D. Teneketzis. Multi-armed bandit problems. In *Foundations and Applications of Sensor Management*, pages 121–151. Springer, 2008.

[20] P. Renner and K. Schmedders. A Polynomial Optimization Approach to Principal-Agent Problems. *Econometrica*, 83(2):729–769, 2015.

[21] S. L. Scott. A modern bayesian look at the multi-armed bandit. *Applied Stochastic Models in Business and Industry*, 26(6):639–658, 2010.

[22] S. Sen, A. Ridgway, and M. Ripley. Adaptive budgeted bandit algorithms for trust development in a supply-chain. In *Proceedings of the 2015 International Conference on Autonomous Agents and Multiagent Systems*, pages 137–144. International Foundation for Autonomous Agents and Multiagent Systems, 2015.

[23] E. Solan and N. Vieille. Quitting games. *Mathematics of Operations Research*, 26(2):265–285, 2001.

[24] E. Solan and N. Vieille. Quitting games - An example. *International Journal of Game Theory*, 31(3):365–381, 2002.

[25] S. Vakili and Q. Zhao. Risk-Averse Multi-Armed Bandit Problems under Mean-Variance Measure. *IEEE Journal on Selected Topics in Signal Processing*, 10(6):1093–1111, 2016.

[26] A. S. Vezhnevets, S. Osindero, T. Schaul, N. Heess, M. Jaderberg, D. Silver, and K. Kavukcuoglu. Feudal networks for hierarchical reinforcement learning. *arXiv preprint arXiv:1703.01161*, 2017.

[27] P. D. Welch. The statistical analysis of simulation results. *The computer performance modeling handbook*, 22:268–328, 1983.

[28] P. Whittle. Sequential Decision Processes with Essential Unobservables. *Advances in Applied Probability Adv. Appl. Prob*, 1(1):271–287, 1969.

[29] P. Whittle. Arm-acquiring bandits. *The Annals of Probability*, 9(2):284–292, 1981.